\documentclass[10pt,twocolumn,letterpaper]{article}

\usepackage{cvpr}
\usepackage{times}
\usepackage{epsfig}
\usepackage{graphicx}
\usepackage{amsmath}
\usepackage{amssymb}
\usepackage{array}
\usepackage{makecell}
\usepackage{multirow}
\usepackage{animate}
\usepackage{wrapfig}
\usepackage[misc]{ifsym} 
\newcolumntype{P}[1]{>{\centering\arraybackslash}p{#1}}

\usepackage[pagebackref=true,breaklinks=true,letterpaper=true,colorlinks,bookmarks=false]{hyperref}

\newcounter{alphasect}
\def\alphainsection{0}

\let\oldsection=\section
\def\section{%
  \ifnum\alphainsection=1%
    \addtocounter{alphasect}{1}
  \fi%
\oldsection}%

\renewcommand\thesection{%
  \ifnum\alphainsection=1%
    \Alph{alphasect}
  \else%
    \arabic{section}
  \fi%
}%

\newenvironment{alphasection}{%
  \ifnum\alphainsection=1%
    \errhelp={Let other blocks end at the beginning of the next block.}
    \errmessage{Nested Alpha section not allowed}
  \fi%
  \setcounter{alphasect}{0}
  \def\alphainsection{1}
}{%
  \setcounter{alphasect}{0}
  \def\alphainsection{0}
}%

\cvprfinalcopy 

\def\cvprPaperID{1791} 

\ifcvprfinal\fi
\begin{document}

\title{Deep Image Spatial Transformation for Person Image Generation}

\author{Yurui Ren$^{1,2}$~~~Xiaoming Yu$^{1,2}$~~~Junming Chen$^{1,2}$~~~Thomas H. Li$^{3,1}$~~~{Ge Li \footnotesize{\Letter}}$^{1,2}$\\
$^1$School of Electronic and Computer Engineering, Peking University~~~$^2$Peng Cheng Laboratory~~~\\
$^3$Advanced Institute of Information Technology, Peking University~~\\
{\tt\small ~~\{yrren,xiaomingyu,junming.chen\}@pku.edu.cn~~~~tli@aiit.org.cn~~~~geli@ece.pku.edu.cn~~~}\\  
}

\maketitle

\begin{abstract}
Pose-guided person image generation is to transform a source person image to a target pose. This task requires spatial manipulations of source data. However, Convolutional Neural Networks are limited by the lack of ability to spatially transform the inputs. In this paper, we propose a differentiable global-flow local-attention framework to reassemble the inputs at the feature level. Specifically, our model first calculates the global correlations between sources and targets to predict flow fields. Then, the flowed local patch pairs are extracted from the feature maps to calculate the local attention coefficients. Finally, we warp the source features using a content-aware sampling method with the obtained local attention coefficients. The results of both subjective and objective experiments demonstrate the superiority of our model. Besides, additional results in video animation and view synthesis show that our model is applicable to other tasks requiring spatial transformation. Our source code is available at \url{https://github.com/RenYurui/Global-Flow-Local-Attention}.

\end{abstract}


\section{Introduction}





\begin{figure}[t]
\begin{center}
\includegraphics[width=1\linewidth]{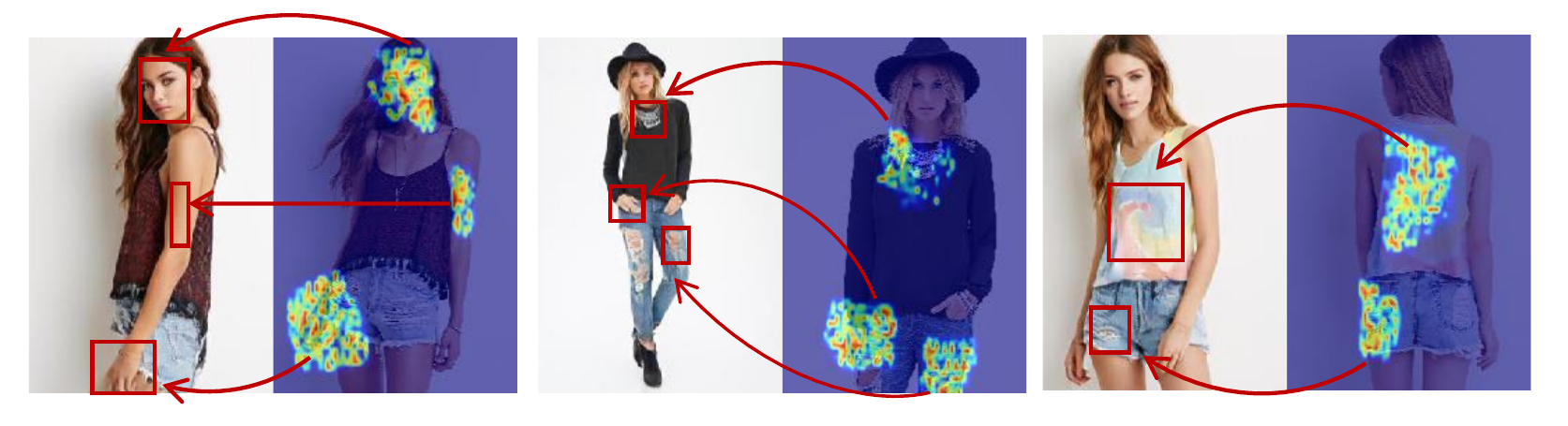}
\end{center}
   \caption{The visualization of data spatial transformation. For each image pair, the left image is the generated result of our model, while the right image is the input source image. Our model spatially transforms the information from sources to targets at the features level. The heat maps indicate the attention coefficients. }
 
\label{fig:intr}
\end{figure}

Image spatial transformation can be used to deal with the generation task where the output images are the spatial deformation versions of the input images. Such deformation can be caused by object motions or viewpoint changes. 
Many conditional image generation tasks can be seen as a type of spatial transformation tasks. For example, pose-guided person image generation~\cite{ma2017pose,siarohin2018deformable,song2019unsupervised,zhu2019progressive,tang2019cycle,tang2020multi} transforms a person image from a source pose to a target pose while retaining the appearance details. As shown in Figure~\ref{fig:intr}, this task can be tackled by reasonably reassembling the input data in the spatial domain.

However, Convolutional Neural Networks (CNNs) are inefficient to spatially transform the inputs. CNNs calculate the outputs with a particular form of parameter sharing, which leads to an important property called equivariance to transformation~\cite{goodfellow2016deep}. It means that if the input spatially shifts the output shifts in the same way. 
This property can benefit tasks 
such as segmentation~\cite{girshick2014rich,he2017mask}, detection~\cite{simonyan2014two,huang2019spatial} and image translation with aligned structures~\cite{isola2017image,yu2019multi} \etc. However, it limits the networks by lacking abilities to spatially rearrange the input data. 
Spatial Transformer Networks (STN)~\cite{jaderberg2015spatial} solves this problem by introducing a Spatial Transformer module to standard neural networks. This module regresses global transformation parameters and warps input features with an affine transformation.  However, since it assumes a global affine transformation between sources and targets, this method cannot deal with the transformations of non-rigid objects.

Attention mechanism~\cite{vaswani2017attention,zhang2018self} allows networks to take use of non-local information, which gives networks abilities to build long-term correlations. It has been proved to be efficient in many tasks such as natural language processing~\cite{vaswani2017attention}, image recognition~\cite{wang2018non,hu2019local}, and image generation~\cite{zhang2018self}. However, for spatial transformation tasks in which target images are the deformation results of source images, each output position has a clear one-to-one relationship with the source positions. Therefore, the attention coefficient matrix between the source and target should be a sparse matrix instead of a dense matrix.  

Flow-based operation forces the attention coefficient matrix to be a sparse matrix by sampling a very local source patch for each output position. 
These methods predict 2-D coordinate offsets specifying which positions in the sources could be sampled to generate the targets. However, in order to stabilize the training, most of the flow-based methods~\cite{zhou2016view,fischer2015flownet} warp input data at the pixel level, which limits the networks to be unable to generate new contents. Meanwhile, large motions are difficult to be extracted due to the requirement of generating full-resolution flow fields~\cite{ranjan2017optical}. Warping the inputs at the feature level can solve these problems. However, the networks are easy to be stuck within bad local minima~\cite{ren2019structureflow,yu2018generative} due to two reasons. $(1)$ The input features and flow fields are mutually constrained. The input features can not obtain reasonable gradients without correct flow fields. The network also cannot extract similarities to generate correct flow fields without reasonable features. $(2)$ The poor gradient propagation provided by the commonly used Bilinear sampling method further lead to instability in training~\cite{jiang2019linearized,ren2019structureflow}. See Section~\ref{sec:Bilinear} for more discussion.

\begin{figure*}[t]
\begin{center}
\includegraphics[width=1\linewidth]{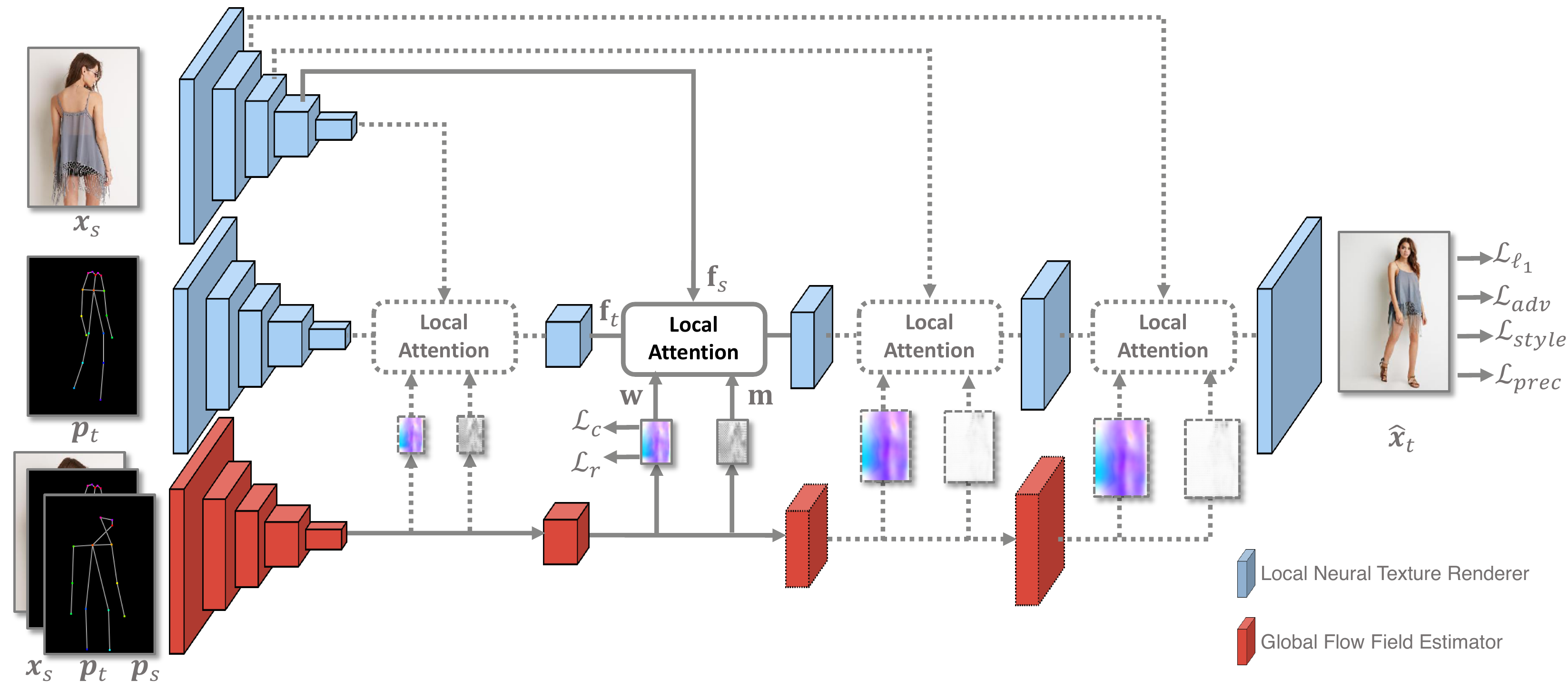}
\end{center}
   \caption{Overview of our model. The Global Flow Field Estimator is used to generate flow fields. The Local Neural Texture Renderer yields results by spatially transforming the source features using local attention. Dotted lines indicate that our local attention module can be used at different scales.}
 
\label{fig:OurModel}
\end{figure*}

In order to deal with these problems, in this paper, we combine flow-based operation with attention mechanisms. We propose a novel global-flow local-attention framework to force each output location to be only related to a local feature patch of sources. The architecture of our model can be found in Figure~\ref{fig:OurModel}.
Specifically, our network can be divided into two parts: Global Flow Field Estimator and Local Neural Texture Renderer. The Global Flow Filed Estimator is responsible for extracting the global correlations and generating flow fields. The Local Neural Texture Renderer is used to sample vivid source textures to targets according to the obtained flow fields. To avoid the poor gradient propagation of the Bilinear sampling, we propose a local attention mechanism as a content-aware sampling method.
We compare our model with several state-of-the-art methods. The results of both subjective and objective experiments show the superior performance of our model. We also conduct comprehensive ablation studies to verify our hypothesis. Besides, we apply our model to other tasks requiring spatial transformation manipulation including view synthesis and video animation. The results show the versatility of our module. The main contributions of our paper can be summarized as: 




\vspace{-3.5mm}
\begin{itemize}
  \setlength\itemsep{-0.9mm}
  \item A global-flow local-attention framework is proposed for the pose-guided person image generation task. Experiments demonstrate the effectiveness of the proposed method.
  \item The carefully-designed framework and content-aware sampling operation ensure that our model is able to warp and reasonably reassemble the input data at the feature level. This operation not only enables the model to generate new contents, but also reduces the difficulty of the flow field estimation task.
  \item Additional experiments on view synthesis and video animation show that our model can be flexibly applied to different tasks requiring spatial transformation.
\end{itemize}
\section{Related Work}

\noindent
\textbf{Pose-guided Person Image Generation.}
An early attempt~\cite{ma2017pose} on the pose-guided person image generation task proposes a two-stage network to first generate a coarse image with target pose and then refine the results in an adversarial way. Essner \etal~\cite{esser2018variational} try to disentangle the appearance and pose of person images. Their model enables both conditional image generation and transformation. However, they use U-Net based skip connections, which may lead to feature misalignments.
Siarohin \etal~\cite{siarohin2018deformable} solve this problem by introducing deformable skip connections to spatially transform the textures. It decomposes the overall deformation by a set of local affine transformations (\eg arms and legs \etc). Although it works well in person image generation, the requirement of the pre-defined transformation components limits its application. Zhu \etal~\cite{zhu2019progressive} propose a more flexible method by using a progressive attention module to transform the source data. However, useful information may be lost during multiple transfers, which may result in blurry details. Han \etal~\cite{han2019clothflow} use a flow-based method to transform the source information. However, they warp the sources at the pixel level, which means that further refinement networks are required to fill the holes of occlusion contents. Liu \etal~\cite{liu2019liquid} and Li \etal~\cite{li2019dense} warp the inputs at the feature level. But both of them need additional 3D human models to calculate the flow fields between sources and targets, which limits the application of these models. Our model does not require any supplementary information and obtains the flow fields in an unsupervised manner.


\noindent
\textbf{Image Spatial Transformation.}
Many methods have been proposed to enable the spatial transformation capability of Convolutional Neural Networks. Jaderberg \etal~\cite{jaderberg2015spatial} introduce a differentiable Spatial Transformer module that estimates global transformation parameters and warps the features with affine transformation. 
Several variants have been proposed to improve the performance.
Zhang \etal add controlling points for free-form deformation~\cite{zhang2017deep}. The model proposed in paper~\cite{lin2017inverse} sends the transformation parameters instead of the transformed features to the network to avoid sampling errors. Jiang \etal~\cite{jiang2019linearized} demonstrate the poor gradient propagation of the commonly used Bilinear sampling. They propose a linearized multi-sampling method for spatial transformation.

Flow-based methods are more flexible than affine transformation methods. They can deal with complex deformations. Appearance flow~\cite{zhou2016view} predicts flow fields and generates the targets by warping the sources. However, it warps image pixels instead of features. This operation limits the model to be unable to generate new contents. Besides, it requires the model to predict flow fields with the same resolution as the result images, which makes it difficult for the model to capture large motions~\cite{liu2017voxelflow,ranjan2017optical}. Vid2vid~\cite{wang2018video} deals with these problems by predicting the ground-truth flow fields using FlowNet~\cite{fischer2015flownet} first and then trains their flow estimator in a supervised manner. They also use a generator for occluded content generation. 
Warping the sources at the feature level can avoid these problems. In order to stabilize the training, some papers propose to obtain the flow-fields by using some assumptions or supplementary information. Paper~\cite{siarohin2019animating} assumes that keypoints are located on object parts that are locally rigid. They generate dense flow fields from sparse keypoints. Papers~\cite{liu2019liquid,li2019dense} use the 3D human models and the visibility maps to calculate the flow fields between sources and targets. Paper~\cite{ren2019structureflow} proposes a sampling correctness loss to constraint flow fields and achieve good results.


\section{Our Approach}
For the pose-guided person image generation task, target images are the deformation results of source images, which means that each position of targets is only related to a local region of sources. Therefore, we design a global-flow local-attention framework to reasonably sample and reassemble source features.
Our network architecture is shown in Figure~\ref{fig:OurModel}. It consists of two modules: \textit{Global Flow Field Estimator} $F$ and \textit{Local Neural Texture Renderer} $G$. The Global Flow Field Estimator is responsible for estimating the motions between sources and targets. It generates global flow fields $\mathbf{w}$ and occlusion masks $\mathbf{m}$ for the local attention blocks. With $\mathbf{w}$ and $\mathbf{m}$, the Local Neural Texture Renderer renders the target images with vivid source features using the local attention blocks. We describe the details of these modules in the following sections. Please note that to simplify the notations, we describe the network with a single local attention block. As shown in Figure~\ref{fig:OurModel}, our model can be extended to use multiple attention blocks at different scales.




\begin{figure*}[t]
\begin{center}
\includegraphics[width=0.95\linewidth]{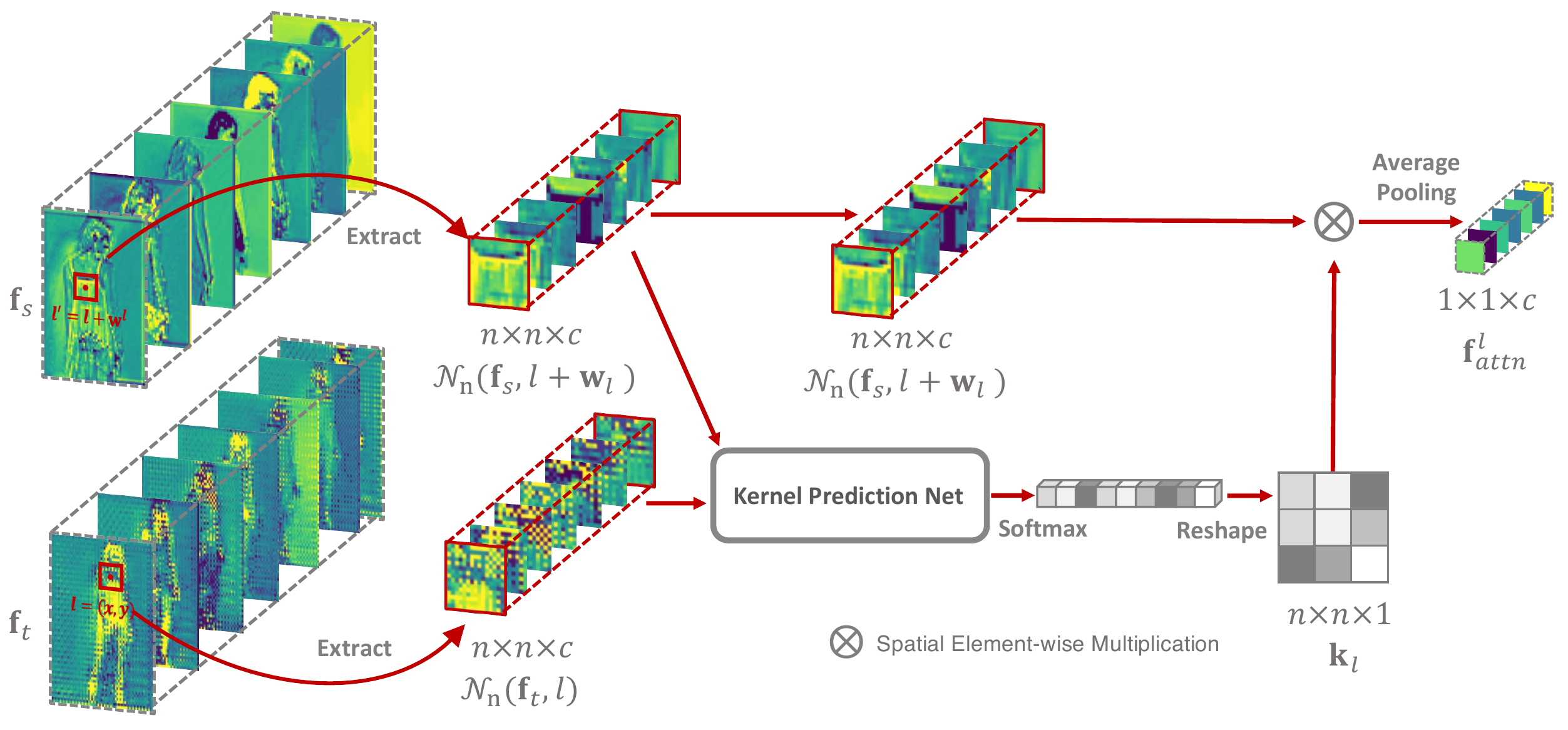}
\end{center}
   \caption{Overview of our Local Attention. We first extract the feature patch pair from the source and target according to the flow fields. Then the context-aware sampling kernel is calculated by the kernel prediction net. Finally, we sample the source feature and obtain the warped result located at $l$.}
 
\label{fig:LocalAttention}
\end{figure*}

\subsection{Global Flow Field Estimator}
Let $\mathbf{p}_s$ and $\mathbf{p}_t$ denote the structure guidance of the source image $\mathbf{x}_s$ and the target image $\mathbf{x}_t$ respectively.
Global Flow Field Estimator $F$ is trained to predict the motions between $\mathbf{x}_s$ and $\mathbf{x}_t$ in an unsupervised manner.  
It takes $\mathbf{x}_s$, $\mathbf{p}_s$ and $\mathbf{p}_t$ as inputs and generates flow fields $\mathbf{w}$ and occlusion masks $\mathbf{m}$.

\begin{equation}
    \mathbf{w},\mathbf{m}=F(\mathbf{x}_s, \mathbf{p}_s, \mathbf{p}_t)
\end{equation}
where $\mathbf{w}$ contains the coordinate offsets between sources and targets. The occlusion mask $\mathbf{m}$ with continuous values between $0$ and $1$ indicates whether the information of a target position exists in the sources. We design $F$ as a fully convolutional network. $\mathbf{w}$ and $\mathbf{m}$ share all weights of $F$ other than their output layers.

As the labels of the flow fields $\mathbf{w}$ are always unavailable in this task, 
we use the sampling correctness loss proposed by~\cite{ren2019structureflow} to constraint $\mathbf{w}$. It calculates the similarity between the warped source feature and ground-truth target feature at the VGG feature level. 
Let $\mathbf{v}_{s}$ and $\mathbf{v}_{t}$ denote the features generated by a specific layer of VGG19. $\mathbf{v}_{s,\mathbf{w}}=\mathbf{w}(\mathbf{v}_{s})$ is the warped results of the source feature $\mathbf{v}_{s}$ using $\mathbf{w}$.
The sampling correctness loss calculates the relative cosine similarity between $\mathbf{v}_{s,\mathbf{w}}$ and $\mathbf{v}_{t}$.
\begin{equation}
   \mathcal{L}_{c} = \frac{1}{N}\sum_{l\in \Omega}exp(-\frac{\mu(\mathbf{v}_{s,\mathbf{w}}^l, \mathbf{v}_{t}^l)}{\mu_{max}^l})
\end{equation}
where $\mu(*)$ denotes the cosine similarity. Coordinate set $\Omega$ contains all $N$ positions in the feature maps, and $\mathbf{v}_{s,\mathbf{w}}^l$ denotes the feature of $\mathbf{v}_{s,\mathbf{w}}$ located at the coordinate $l=(x, y)$. The normalization term $\mu_{max}^l$ is calculated as
\begin{equation}
   \mu_{max}^l = \max\limits_{l'\in \Omega}\mu(\mathbf{v}_{s}^{l'}, \mathbf{v}_{t}^{l})
\end{equation}
It is used to avoid the bias brought by occlusion.

The sampling correctness loss can constrain the flow fields to sample semantically similar regions. However, as the deformations of image neighborhoods are highly correlated, it would  benefit if we could extract this relationship. Therefore, we further add a regularization term to our flow fields. This regularization term is used to punish local regions where the transformation is not an affine transformation. Let $\mathbf{c}_{t}$ be the 2D coordinate matrix of the target feature map. The corresponding source coordinate matrix can be written as $\mathbf{c}_{s} = \mathbf{c}_{t}+\mathbf{w}$. We use $\mathcal{N}_n(\mathbf{c}_{t}, l)$ to denote local $n\times n$ patch of $\mathbf{c}_{t}$ centered at the location $l$.
Our regularization assumes that the transformation between $\mathcal{N}_n(\mathbf{c}_{t}, l)$ and $\mathcal{N}_n(\mathbf{c}_{s}, l)$ is an affine transformation.
\begin{equation}
    \mathbf{T}_l={\mathbf{A}}_l\mathbf{S}_l=\left[ {\begin{array}{ccc}\theta_{11}&\theta_{12}&\theta_{13}\\
    \theta_{21}&\theta_{22}&\theta_{23}\end{array}} \right]\mathbf{S}_l
\end{equation}
where $\mathbf{T}_l=\left[{\begin{array}{cccc}x_1&x_2&...&x_{n\times n}\\
    y_1&y_2&...&y_{n \times n}\end{array}}\right]$ with each coordinate $(x_i,y_i)\in \mathcal{N}_n(\mathbf{c}_t, l)$ and $\mathbf{S}_l=\left[{\begin{array}{cccc}x_1&x_2&...&x_{n\times n}\\
    y_1&y_2&...&y_{n\times n}\\1&1&...&1\end{array}}\right]$ with each coordinate $(x_i,y_i)\in \mathcal{N}_n(\mathbf{c}_s, l)$.
    The estimated affine transformation parameters $\hat{\mathbf{A}}_l$ can be solved using the least-squares estimation as  
\begin{equation}
    \hat{\mathbf{A}}_l = (\mathbf{S}_l^H\mathbf{S}_l)^{-1}\mathbf{S}_l^H\mathbf{T}_l
\end{equation}
Our regularization is calculated as the $\ell_2$ distance of the error.
\begin{equation}
    \mathcal{L}_r=\sum_{l \in \Omega} \left\lVert \mathbf{T}_l - \hat{\mathbf{A}}_l{\mathbf{S}}_l \right\lVert_2^2
\end{equation}

\begin{table*}
\setlength\extrarowheight{1pt}
\centering
\resizebox{.9\textwidth}{!}{%

\begin{tabular}{p{2cm}||P{1.2cm}P{1.2cm}P{1.2cm}||P{1.2cm}P{1.2cm}P{2cm}P{1.2cm}||P{2cm}}
\hline
          & \multicolumn{3}{c||}{DeepFashion} & \multicolumn{4}{c||}{Market-1501}&  Number of \\ \cline{2-8} 
          & FID        & LPIPS      & JND    & FID    & LPIPS  & Mask-LPIPS & JND & Parameters\\ \hline
          &            &            &        &        &        &            & & \\ [-11pt]\hline
Def-GAN   & 18.457     & 0.2330     & 9.12\%     & 25.364 & 0.2994 & 0.1496     &23.33\%     & 82.08M\\ \hline
VU-Net    & 23.667     & 0.2637     & 2.96\% & 20.144 & 0.3211 & 0.1747     & 24.48\% & 139.36M\\ \hline
Pose-Attn & 20.739     & 0.2533     & 6.11\% & 22.657 & 0.3196 & 0.1590     & 16.56\%  & 41.36M\\ \hline
Intr-Flow & 16.314     & \textbf{0.2131} & 12.61\%  & 27.163 & 0.2888 & \textbf{0.1403}     & \textbf{30.85}\%      &    49.58M      \\ \hline
Ours      & \textbf{10.573}& 0.2341 &\textbf{24.80}\%      & \textbf{19.751} & \textbf{0.2817} & 0.1482&27.81\%    & \textbf{14.04M}\\ \hline
\end{tabular}%

}
\caption{The evaluation results compared with several state-of-the-art methods including Def-GAN~\cite{siarohin2018deformable}, VU-Net~\cite{esser2018variational}, Pose-Attn\cite{zhu2019progressive}, and Intr-Flow~\cite{li2019dense} over dataset DeepFashion~\cite{liu2016deepfashion} and Market-1501~\cite{zheng2015scalable}. FID~\cite{heusel2017gans} and LPIPS~\cite{zhang2018unreasonable} are objective metrics. JND is obtained by human subjective studies. It represents the probability that the generated images are mistaken for real images.}
\label{tab:object}
\end{table*}

\subsection{Local Neural Texture Renderer}
\label{TargetImageGenerator}
With the flow fields $\mathbf{w}$ and occlusion masks $\mathbf{m}$, our Local Neural Texture Renderer $G$ is responsible for generating the results by spatially transforming the information from sources to targets.
It takes $\mathbf{x}_s$, $\mathbf{p}_t$, $\mathbf{w}$ and $\mathbf{m}$ as inputs and generate the result image $\hat{\mathbf{x}}_t$.
\begin{equation}
    \hat{\mathbf{x}}_t=G(\mathbf{x}_s, \mathbf{p}_t, \mathbf{w},\mathbf{m})
\end{equation}

Specifically, the information transformation occurs in the local attention module.
As shown in Figure~\ref{fig:OurModel}, this module works as a neural renderer where the target bones are rendered by the neural textures of the sources. Let $\mathbf{f}_t$ and $\mathbf{f}_s$ represent the extracted features of target bones $\mathbf{p}_t$ and source images $\mathbf{x}_s$ respectively.
We first extract local patches $\mathcal{N}_n(\mathbf{f}_t, l)$ and $\mathcal{N}_n(\mathbf{f}_s, l+\mathbf{w}^l)$ from $\mathbf{f}_t$ and $\mathbf{f}_s$ respectively. The patch $\mathcal{N}_n(\mathbf{f}_s, l+\mathbf{w}^l)$ is extracted using bilinear sampling as the coordinates may not be integers.
Then, a kernel prediction network $M$ is used to predict local $n \times n$ kernel $\mathbf{k}_l$ as 
\begin{equation}
    \mathbf{k}_l=M(\mathcal{N}_n(\mathbf{f}_s, l+\mathbf{w}^l), \mathcal{N}_n(\mathbf{f}_t, l))
\end{equation}
We design $M$ as a fully connected network, where the local patches $\mathcal{N}_n(\mathbf{f}_s, l+\mathbf{w}^l)$ and $\mathcal{N}_n(\mathbf{f}_t, l)$ are directly concatenated as the inputs.
The softmax function is used as the non-linear activation function of the output layer of $M$.
This operation forces the sum of $\mathbf{k}_l$ to 1, which enables the stability of gradient backward. Finally, the flowed feature located at coordinate $l=(x,y)$ is calculated using a content-aware attention over the extracted source feature patch $\mathcal{N}_n(\mathbf{f}_s,l+\mathbf{w}^l)$.
\begin{equation}
    \mathbf{f}^l_{attn}=P(\mathbf{k}_l\otimes\mathcal{N}_n(\mathbf{f}_s, l+\mathbf{w}^l))
\end{equation}
where $\otimes$ denotes the element-wise multiplication over the spatial domain and $P$ represents the global average pooling operation. The warped feature map $\mathbf{f}_{attn}$ is  obtained by repeating the previous steps for each location $l$. 

However, not all contents of target images can be found in source images because of occlusion or movements. In order to enable generating new contents, the occlusion mask $\mathbf{m}$ with continuous value between 0 and 1 is used to select features between $\mathbf{f}_{attn}$ and $\mathbf{f}_{t}$.
\begin{equation}
    \mathbf{f}_{out} = (\mathbf{1}-\mathbf{m})*\mathbf{f}_t + \mathbf{m}*\mathbf{f}_{attn}
\end{equation}

We train the network using a joint loss consisting of a reconstruction $\ell_1$ loss, adversarial loss, perceptual loss, and style loss.
The reconstruction $\ell_1$ loss is written as 
\begin{equation}
    \mathcal{L}_{\ell_1}=\left\lVert \mathbf{x}_t - \hat{\mathbf{x}}_t \right\lVert_1
\end{equation}
The generative adversarial framework~\cite{goodfellow2014generative} is employed to mimic the distributions of the ground-truth $\mathbf{x}_{t}$. The adversarial loss is written as 
\begin{align}
   \mathcal{L}_{adv} &= \mathbb{E}[\log(1-D(G(\mathbf{x}_s, \mathbf{p}_t, \mathbf{w},\mathbf{m})))] \notag \\
   &+ \mathbb{E}[\log D(\mathbf{x}_{t})]
   \label{eq:advr}   
\end{align}
where $D$ is the discriminator of the Local Neural Texture Renderer $G$. 
We also use the perceptual loss and style loss introduced by~\cite{johnson2016perceptual}. The perceptual loss calculates $\ell_1$ distance between activation maps of a pre-trained network. It can be written as
\begin{equation}
    \mathcal{L}_{perc}=\sum_i\left\lVert \phi_i(\mathbf{x}_t) - \phi_i(\hat{\mathbf{x}}_t)\right\lVert_1
\end{equation}
where $\phi_i$ is the activation map of the $i$-th layer of a pre-trained network. The style loss calculates the statistic error between the  activation maps as
\begin{equation}
    \mathcal{L}_{style}=\sum_j\left\lVert G_j^\phi(\mathbf{x}_t) - G^\phi_j(\hat{\mathbf{x}}_t)\right\lVert_1
\end{equation}
where $G_j^\phi$ is the Gram matrix constructed from activation maps $\phi_j$.
We train our model using the overall loss as 
\begin{equation}
    \mathcal{L}=\lambda_c\mathcal{L}_{c} + \lambda_r\mathcal{L}_{r} + \lambda_{\ell_1}\mathcal{L}_{\ell_1}+ \lambda_a\mathcal{L}_{adv}+ \lambda_p\mathcal{L}_{prec}+ \lambda_s\mathcal{L}_{style}
\end{equation}

\section{Experiments}

\subsection{Implementation Details}
\noindent
\textbf{Datasets.} Two datasets are used in our experiments: person re-identification dataset Market-1501~\cite{zheng2015scalable} and DeepFashion In-shop Clothes Retrieval Benchmark~\cite{liu2016deepfashion}. Market-1501 contains 32668 low-resolution images (128 $\times$ 64). The images vary in terms of the viewpoints, background, illumination \etc. The DeepFashion dataset contains 52712 high-quality model images with clean backgrounds. We split the datasets with the same method as that of~\cite{zhu2019progressive}. The personal identities of the training and testing sets do not overlap.

\begin{figure*}[t]
\begin{center}
\includegraphics[width=1\linewidth]{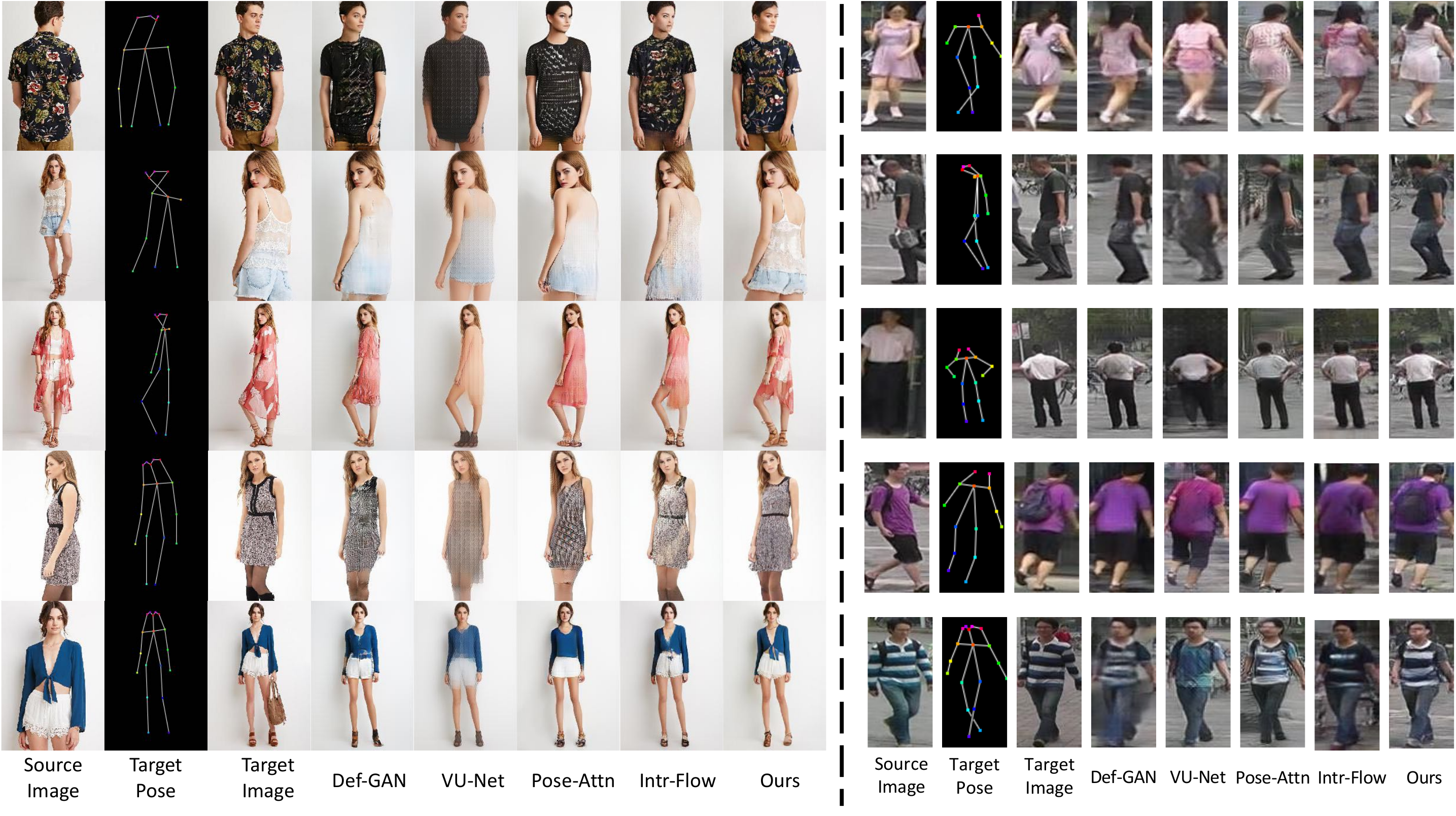}
\end{center}
\caption{The qualitative comparisons with several state-of-the-art models including Def-GAN~\cite{siarohin2018deformable}, VU-Net~\cite{esser2018variational}, Pose-Attn\cite{zhu2019progressive}, and Intr-Flow~\cite{li2019dense}. The left part shows the results of the Fashion dataset. The right part shows the results of the Market-1501 dataset.}
\label{fig:Compare}

\end{figure*}

\noindent
\textbf{Metrics.} We use Learned Perceptual Image Patch Similarity (LPIPS) proposed by~\cite{zhang2018unreasonable} to calculate the reconstruction error. LPIPS computes the distance between the generated images and reference images at the perceptual domain. It indicates the perceptual difference between the inputs. Meanwhile, Fr\'echet Inception Distance~\cite{heusel2017gans} (FID) is employed to measure the realism of the generated images. 
It calculates the Wasserstein-2 distance between distributions of the generated images and ground-truth images. 
Besides, we perform a Just Noticeable Difference (JND) test to evaluate the subjective quality. Volunteers are asked to choose the more realistic image from the data pair of ground-truth and generated images. 


\noindent
\textbf{Network Implementation and Training Details.}
Basically, auto-encoder structures are employed to design our $F$ and $G$. 
The residual block is used as the basic component of these models. 
We train our model using $256 \times 256$ images for the Fashion dataset. Two local attention blocks are used for feature maps with resolutions as $32 \times 32$ and $64 \times 64$. The extracted local patch sizes are $3$ and $5$ respectively. For Market-1501, we use $128 \times 64$ images with a single local attention block at the feature maps with resolution as $32 \times 16$. The extracted patch size is $3$. We train our model in stages. The Flow Field Estimator is first trained to generate flow fields. Then we train the whole model in an end-to-end manner. We adopt the ADAM optimizer with the learning rate as $10^{-4}$. The batch size is set to $8$ for all experiments.

\subsection{Comparisons}
We compare our method with several stare-of-the-art methods including Def-GAN~\cite{siarohin2018deformable}, VU-Net~\cite{esser2018variational}, Pose-Attn\cite{zhu2019progressive} and Intr-Flow~\cite{li2019dense}. 
The quantitative evaluation results are shown in Table~\ref{tab:object}. 
For the Market-1501 dataset, we follow the previous work~\cite{ma2017pose} to calculate the mask-LPIPS to alleviate the influence of the backgrounds. It can be seen that our model achieves competitive results in both datasets, which means that our model can generate realistic results with fewer perceptual reconstruction errors.


As the subjective metrics may not be sensitive to some artifacts, its results may mismatch with the actual subjective perceptions. Therefore, we implement a just noticeable difference test on Amazon Mechanical Turk (MTurk). This experiment requires volunteers to choose the more realistic image from image pairs of real and generated images. The test is performed over $800$ images for each model and dataset. 
Each image is compared $5$ times by different volunteers. The evaluation results are shown in Table~\ref{tab:object}. It can be seen that our model achieves the best result in the challenging Fashion dataset and competitive results in the Market-1501 dataset.

The typical results of different methods are provided in Figure~\ref{fig:Compare}. For the Fashion dataset, VU-Net and Pose-Attn struggle to generate complex textures since these models lack efficient spatial transformation blocks. Def-GAN defines local affine transformation components (\eg arms and legs \etc). This model can generate correct textures. However, the pre-defined affine transformations are not sufficient to represent complex spatial variance, which limits the performance of the model. Flow-based model Intr-Flow is able to generate vivid textures for front pose images. However, it may fail to generate realistic results for side pose images due to the requirement of generating full-resolution flow fields. Meanwhile, this model needs 3D human models to generate the ground-truth flow fields for training. Our model regresses flow fields in an unsupervised manner. It can generate realistic images with not only the correct global pattern but also the vivid details such as the lace of clothes and the shoelace. 
For the Market-1501 Dataset, our model can generate correct pose with vivid backgrounds. Artifacts can be found in the results of competitors, such as the sharp edges in Pose-Attn and the halo effects in Def-GAN. Please refer to Section~\ref{sec:additional_result} for more comparison results.


The numbers of model parameters are also provided to evaluate the computation complexity in Table~\ref{tab:object}. Thanks to our efficient attention blocks, our model does not require a large number of convolution layers. Thus, we can achieve high performance with less than half of the parameters of the competitors.    

\begin{table}[]
\centering
\setlength\extrarowheight{1pt}
\resizebox{.5\textwidth}{!}{%
\centering
\begin{tabular}{c||c|c|P{2cm}|P{2cm}}
\hline
                        &  \multirow{2}{*}{Flow-Based} & Content-aware & \multirow{2}{*}{FID}    & \multirow{2}{*}{LPIPS} \\ 
                        &&Sampling&&  \\\hline
                        &&&&\\ [-11pt]\hline
Baseline                & N          & -                      & 16.008 & 0.2473 \\
Global-Attn    & N          & -                      & 18.616 & 0.2575 \\
Bi-Sample & Y          & N                      & 12.143 & 0.2406 \\
Full Model             & Y          & Y                      & \textbf{10.573} & \textbf{0.2341} \\ \hline
\end{tabular}%
}
\caption{The evaluation results of the ablation study.}
\label{tab:AblationStudy}
\end{table}

\subsection{Ablation Study}
In this subsection, we train several ablation models to verify our assumptions and evaluate the contribution of each component.

\noindent
\textbf{Baseline.}
Our baseline model is an auto-encoder convolutional network. We do not use any attention blocks in this model. Images $\mathbf{x}_s$, $\mathbf{p}_t$, $\mathbf{p}_s$ are directly concatenated as the model inputs.

\noindent
\textbf{Global Attention Model (Global-Attn).}
The Global-Attn model is designed to compare the global-attention block with our local-attention block. We use a similar network architecture as our Local Neural Texture Renderer $G$ for this model. The local attention blocks are replaced by global attention blocks where the attention coefficients are calculated by the similarities between the source features $\mathbf{f}_s$ and target features $\mathbf{f}_t$. 

\noindent
\textbf{Bilinear Sampling Model (Bi-Sample).}
The Bi-Sample model is designed to evaluate the contribution of our content-aware sampling method described in Section~\ref{TargetImageGenerator}. 
Both the Global Flow Field Estimator $F$ and Local Neural Texture Renderer $G$ are employed in this model. However, we use the Bilinear sampling as the sampling method in model $G$.

\noindent
\textbf{Full Model (Ours).}
We use our proposed  global-flow local-attention framework in this model.

\begin{figure}[t]
\begin{center}
\includegraphics[width=1\linewidth]{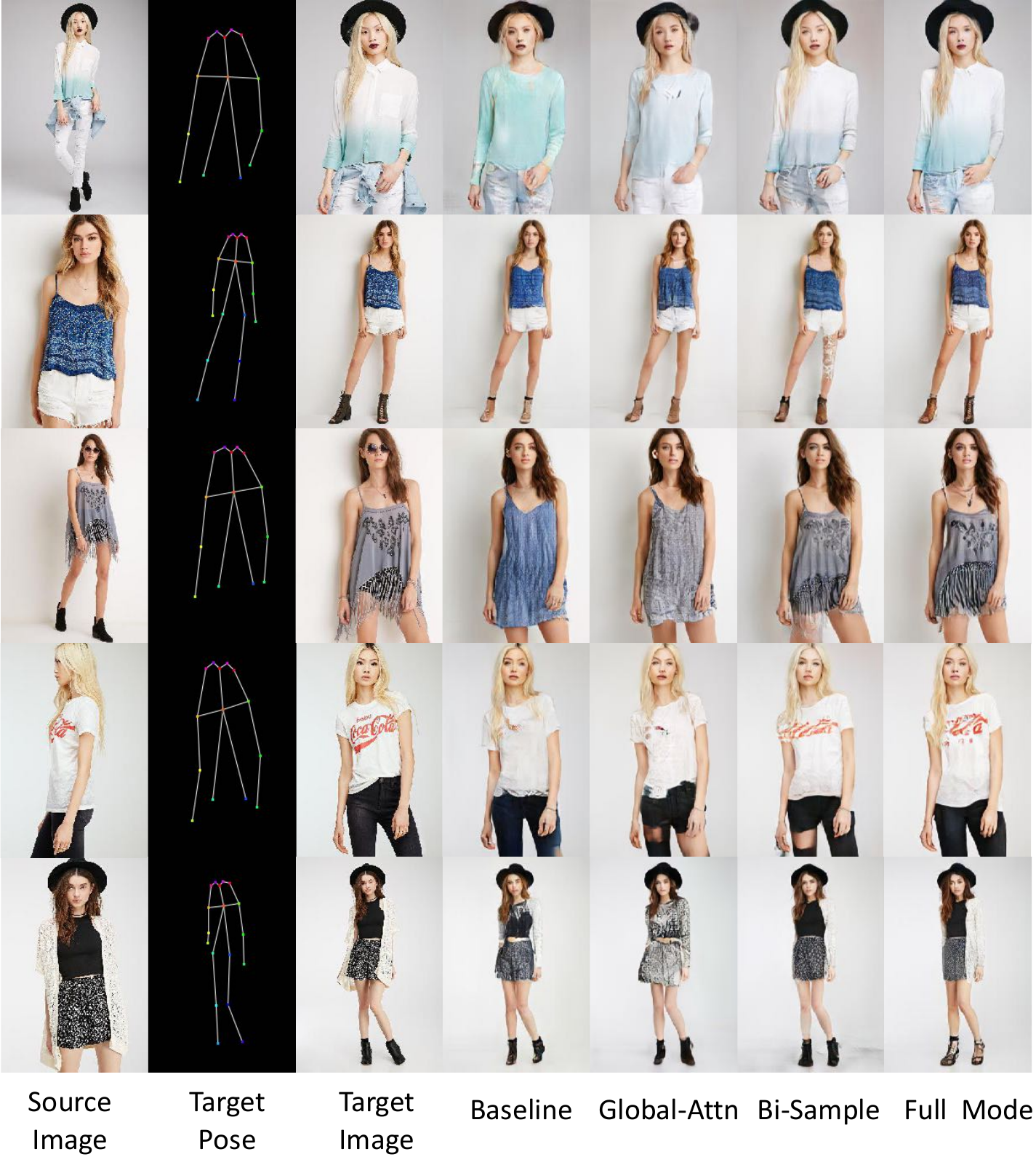}
\end{center}
   \caption{Qualitative results of the ablation study.}
 
\label{fig:ablation}
\end{figure}

\begin{figure}[t]
\begin{center}
\includegraphics[width=1\linewidth]{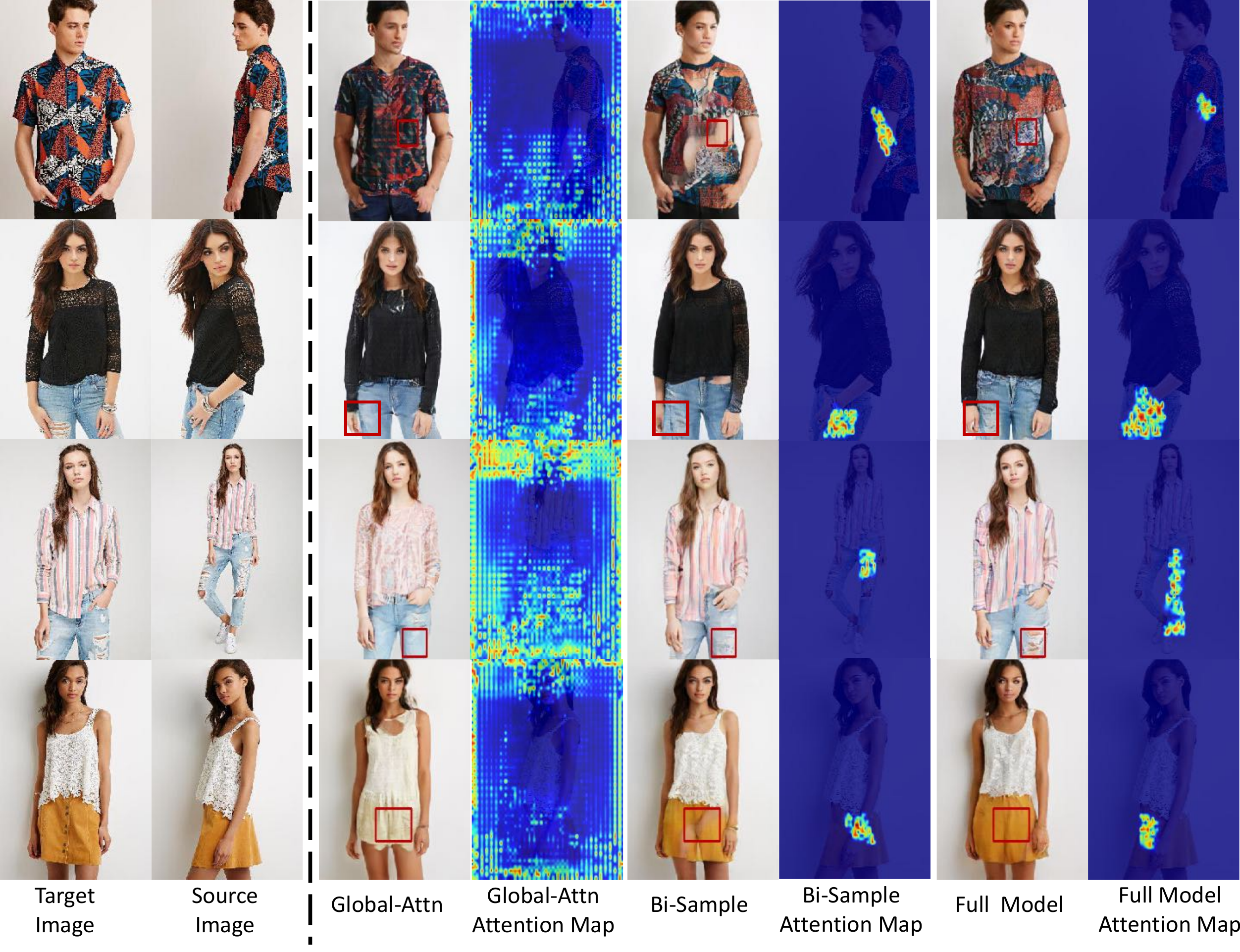}
\end{center}
   \caption{The visualization results of different attention modules. The red rectangles indicate the target locations. The heat maps show the attention coefficients. Blue represents low weights.}
 
\label{fig:ablation_visi_attn}
\end{figure}

The evaluation results of the ablation study are shown in Table~\ref{tab:AblationStudy}. Compared with the Baseline, the performance of the Global-Attn model is degraded, which means that unreasonable attention block cannot efficiently transform the information.
Improvements can be obtained by using flow-based methods such as the Bi-Sample model and our Full model which force the attention coefficient matrix to be a sparse matrix.
However, the Bi-Sample model uses a pre-defined sampling method with a limited sampling receptive field, which may lead to unstable training. Our full model uses a content-aware sampling operation with an adjustable receptive field, which brings further performance gain.

Subjective comparison of these ablation models can be found in Figure~\ref{fig:ablation}. It can be seen that the Baseline and Global-Attn model generate correct structures. However, the textures of the source images are not well-maintained.
The possible explanation is that these models generate images by first extracting global features and then propagating the information to specific locations. This process leads to the loss of details. The flow-based methods spatially transform the features. They are able to reconstruct vivid details. However, the Bi-Sample model uses the pre-defined Bilinear sampling method. It cannot find the exact sampling locations, which leads to artifacts in the final results. 

We further provide the visualization of the attention maps in Figure~\ref{fig:ablation_visi_attn}. It can be seen that the Global-Attn model struggles to exclude irrelevant information. Therefore, the extracted features are hard to be used to generate specific textures. 
The Bi-Sample model assigns a local patch for each generated location. However, incorrect features are often flowed due to the limited sampling receptive field. Our Full model using the content-aware sampling method can flexibly change the sampling weights and avoid the artifacts.



\section{Application on Other Tasks} 
\label{sec:application_on_other_tasks}
In this section, we demonstrate the versatility of our global-flow local-attention module.
Since our model does not require any additional information other than images and structure guidance, it can be flexibly applied to tasks requiring spatial transformation. Two example tasks are shown as follows.



\begin{figure}
    \offinterlineskip
    \centering
    \href{https://user-images.githubusercontent.com/30292465/75651034-81bf4d00-5c92-11ea-93cb-612969b1556a.gif}{
    \includegraphics[width=1\linewidth]{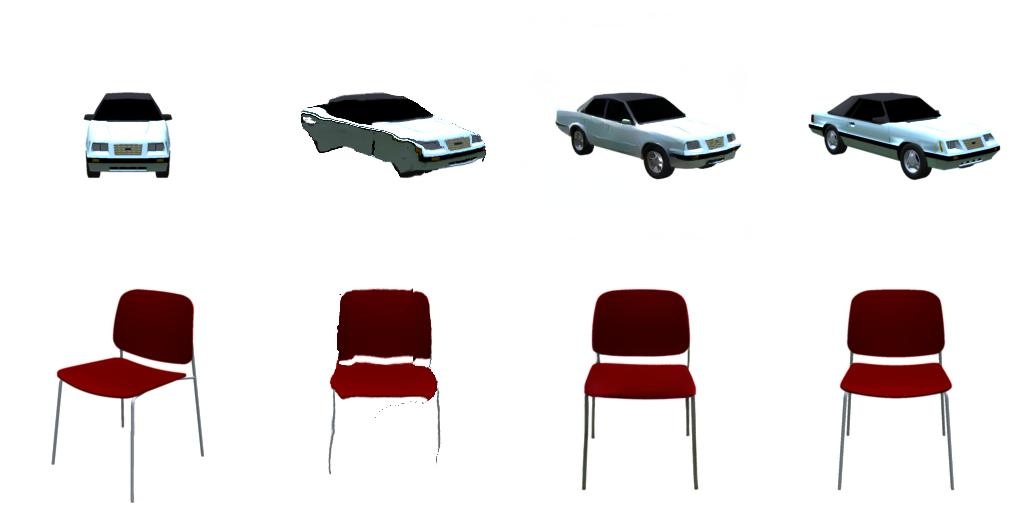}}
    \small
    Source \ \ \ \ \ \ \ \ \ \ \ \ \ \ AppFlow  \ \ \ \ \ \ \ \ \ \ \ \ \  Ours \ \ \ \ \ \ \ \ Ground-Truth 

    \caption{Qualitative results of the view synthesis task. We show the results of our model and appearance flow~\cite{zhou2016view} model. \textit{Click on the image to start the animation in a browser}.}
    \label{fig:view_synthesis}
\end{figure}

\noindent
\textbf{View Synthesis.} 
View synthesis requires generating novel views of objects or scenes based on arbitrary input views. Since the appearance of different views is highly correlated, the existing information can be reassembled to generate the targets. The ShapeNet dataset~\cite{chang2015shapenet} is used for training. We generate novel target views using single view input. The results can be found in Figure~\ref{fig:view_synthesis}. We provide the results of appearance flow as a comparison. It can be seen that appearance flow struggles to generate occluded contents as they warp image pixels instead of features. Our model generates reasonable results.  

\begin{figure}
    \centering
    \href{https://user-images.githubusercontent.com/30292465/75650912-43c22900-5c92-11ea-93b2-f901a5c1fa29.gif}{
    \includegraphics[width=1\linewidth]{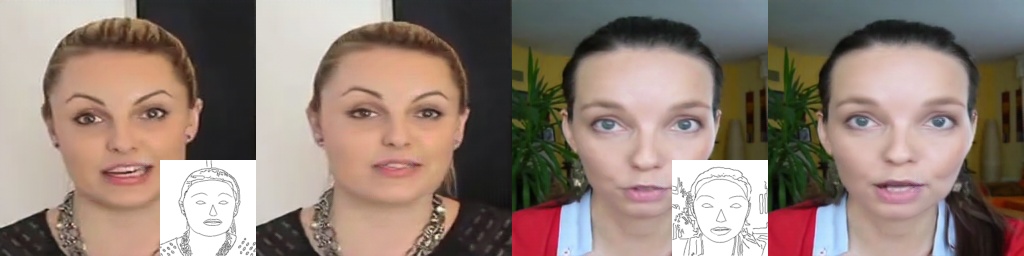}}
    \small
    Source \ \ \ \ \ \ \ \ \ \ \ \ \ \ Result \ \ \ \ \ \ \ \ \ \ \ \ \ \ Source \ \ \ \ \ \ \ \ \ \ \ \ \ \ Result    
    \caption{Qualitative results of the image animation task. Our model generates the results video using reference image and edge guidance. \textit{Click on the image to play the video clip in a browser}.}
    \label{video_animation}
\end{figure}

\noindent
\textbf{Image Animation.} 
Given an input image and a driving video sequence depicting the structure movements, the image animation task requires generating a video containing the specific movements. This task can be solved by spatially moving the appearance of the sources. 
We train our model with the real videos in the FaceForensics dataset~\cite{rossler2018faceforensics}, which contains 1000 videos of news briefings from different reporters. The face regions are cropped for this task. We use the edge maps as the structure guidance. For each frame, the input source frame and the previous generated $n$ frames are used as the references. The flow fields are calculated for each reference. The results can be found in Figure~\ref{video_animation}. It can be seen that our model generates realistic results with vivid movements. More applications can be found in Section~\ref{sec:additional_result}.




\section{Conclusion} 
\label{sec:Conclusion}
In this paper, we solve the person image generation task with deep spatial transformation. We analyze the specific reasons causing instable training when warping and transforming sources at the feature level. Targeted solution global-flow local-attention framework is proposed to enable our model to reasonably reassemble the source neural textures. Experiments show that our model can generate target images with correct poses while maintaining vivid details. In addition, the ablation study shows that our improvements help the network find reasonable sampling positions. Finally, we show that our model can be easily extended to address other spatial deformation tasks such as view synthesis and video animation.


{\small
\bibliographystyle{ieee_fullname}
\bibliography{egbib}
}
\clearpage
\onecolumn
\begin{alphasection}

\section{Warping at the Feature Level}
\label{sec:Bilinear}
The networks are easy to be stuck within bad local minima when warping the inputs at the feature level by using flow-based methods with traditional warping operation. Two main problems are pointed out in our paper. In this section, we further explain these reasons.

Figure~\ref{fig:Bilinear} shows the warping process of the traditional Bilinear sampling. For each location $l=(x,y)$ in the output features $\mathbf{f}_{out}$, a sampling position is assigned by the offsets in flow fields $\mathbf{w}^l=(\Delta{x}, \Delta{y})$. Then, the output feature $\mathbf{f}_{out}^{(x,y)}$ is obtained by sampling the local regions of the input features $\mathbf{f}_{in}$. 


\begin{equation} 
\renewcommand{\theequation}{\Alph{alphasect}.\arabic{equation}}
\begin{split}
\mathbf{f}_{out}^{x,y}& =(\lceil \Delta{y} \rceil-\Delta{y})(\lceil \Delta{x} \rceil-\Delta{x})\mathbf{f}_{in}^{x^{\prime},y^{\prime}} + (\Delta{y}-\lfloor \Delta{y} \rfloor)(\Delta{x}-\lfloor \Delta{x} \rfloor)\mathbf{f}_{in}^{x^{\prime}+1,y^{\prime}+1}\\
& + (\Delta{y}-\lfloor \Delta{y} \rfloor)(\lceil \Delta{x} \rceil-\Delta{x})\mathbf{f}_{in}^{x^{\prime},y^{\prime}+1} + (\lceil \Delta{y} \rceil-\Delta{y})(\Delta{x}-\lfloor \Delta{x} \rfloor)\mathbf{f}_{in}^{x^{\prime}+1,y^{\prime}}
\end{split}
\end{equation}
where $\lceil \cdot \rceil$ and $\lfloor \cdot \rfloor$ represent round up and round down respectively. Location $(x^\prime,y^\prime)=(x+\lfloor x \rfloor,y+\lfloor y \rfloor)$. 
Then, we can obtain the backward gradients. The gradients of flow fields is
\begin{equation}
\renewcommand{\theequation}{\Alph{alphasect}.\arabic{equation}}
\begin{split}
   \frac{\partial{\mathbf{f}^{x,y}_{out}}}{\partial{\Delta x}} = &(\lceil \Delta{y}\rceil-\Delta{y})(\mathbf{f}^{x^{\prime}+1, y^\prime}_{in}-\mathbf{f}^{x^\prime,y^{\prime}}_{in}) +(\Delta{y}-\lfloor \Delta{y}\rfloor)(\mathbf{f}^{x^{\prime}+1, y^\prime+1}_{in}-\mathbf{f}^{x^\prime,y^{\prime}+1}_{in}) 
\end{split}
\label{eq:flow_field}
\end{equation}
the $\frac{\partial{\mathbf{f}^{x,y}_{out}}}{\partial{\Delta y}}$ can be obtained in a similar way. The gradients of the input features can be written as
\begin{equation}
\renewcommand{\theequation}{\Alph{alphasect}.\arabic{equation}}
   \frac{\partial{\mathbf{f}^{x,y}_{out}}}{\partial{\mathbf{f}_{in}^{x^\prime,y^\prime}}}=(\lceil \Delta{y} \rceil-\Delta{y})(\lceil \Delta{x} \rceil-\Delta{x})
\label{eq:inputs}
\end{equation}
The other items can be obtained in a similar way.

\begin{figure}[b]
\renewcommand\thefigure{\Alph{alphasect}.\arabic{figure}}
\begin{center}
\includegraphics[width=0.5\linewidth]{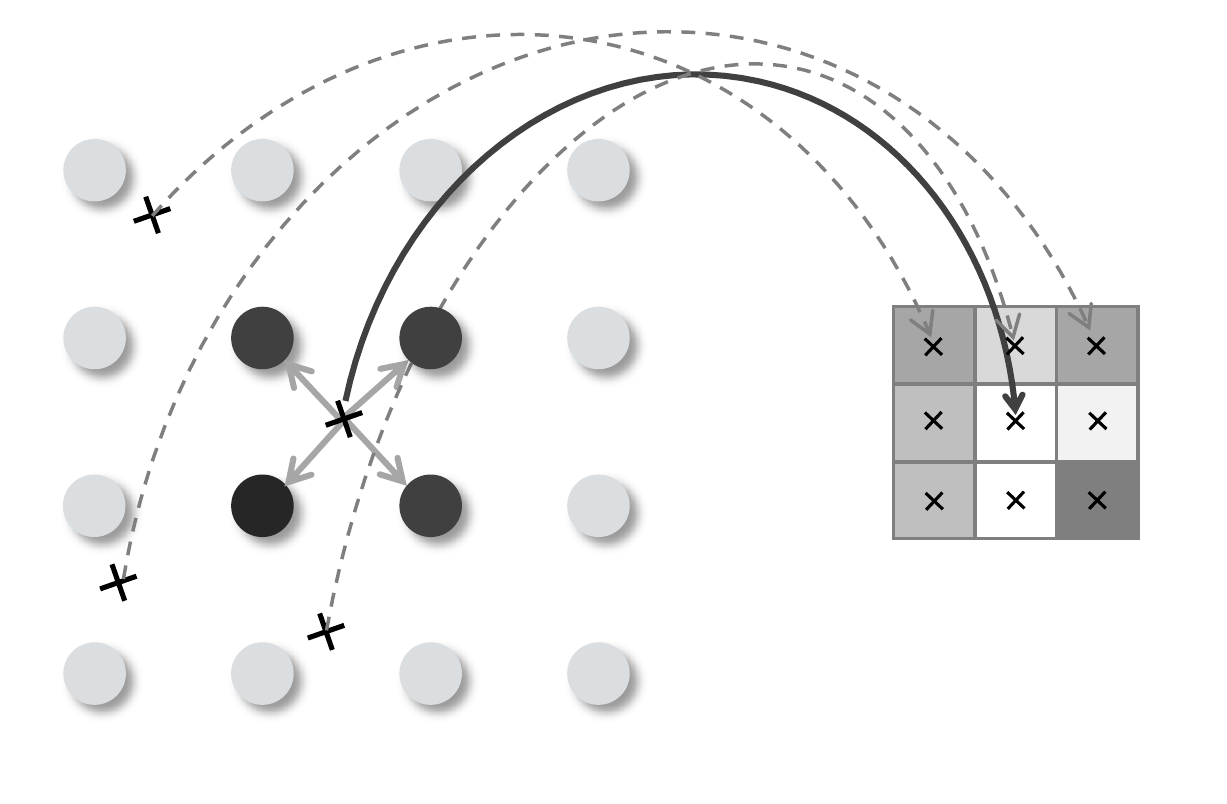}
\end{center}
\caption{Bilinear sampling. For each point of the right feature map (target feature map), the flow field assigns a sampling location in the left feature map (source feature map). Then, the sampled feature is calculated through the nearest neighbors by Bilinear interpolation.}

\label{fig:Bilinear}

\end{figure}


Let us first explain the first problem. Different from image pixels, the image features are changing during the training process. The flow fields need reasonable input features to obtain correct gradients. As shown in Equation~\ref{eq:flow_field}, their gradients are calculated by the difference between adjacent features. If the input features are meaningless, the network cannot obtain correct flow fields. Meanwhile, according to Equation~\ref{eq:inputs}, the gradients of the input features are calculated by the offsets. They cannot obtain reasonable gradients without correct flow fields.
Imagine the worst case that the model only uses the warped results as output and does not use any skip connections, the warp operation stops the gradient from back-propagating.

Another problem is caused by the limited receptive field of Bilinear sampling. Suppose we have got meaningful input features $\mathbf{f}_{in}$ by pre-training, we still cannot obtain stable gradient propagation. According to Equation~\ref{eq:flow_field}, the gradient of flow fields is calculated by the difference between adjacent features. However, since the adjacent features are extracted from adjacent image patches, they often have strong correlations (\ie $\mathbf{f}_{in}^{x^{\prime},y^{\prime}} \approx \mathbf{f}_{in}^{x^{\prime},y^{\prime}+1}$). Therefore, the gradients may be small at most positions. Meanwhile, large motions are hard to be captured. 
\section{Analysis of the Regularization loss}

In our paper, we mentioned that the deformations of image neighborhoods are highly correlated and proposed a regularization loss to extract this relationship. In this section, we further discuss this regularization loss. Our loss is based on an assumption: although the deformations of the whole images are complex, the deformations of local regions such as arms, clothes are always simple. These deformations can be modeled using affine transformation. Based on this assumption, our regularization loss is proposed to punish local regions where the transformation is not an affine transformation. Figure~\ref{fig:regularization} gives an example. Our regularization loss first extracts local 2D coordinate matrix $\mathcal{N}(\mathbf{c}_t, l)$ and $\mathcal{N}(\mathbf{c}_s, l)$. Then we estimate the affine transformation parameters $\hat{\mathbf{A}}_l$. Finally, the error is calculated as the regularization loss.
Our loss assigns large errors to local regions that do not conform to the affine transformation assumption thereby forcing the network to change the sampling regions. 



We provide an ablation study to show the effect of the regularization loss. A model is trained without using the regularization term. The results are shown in Figure~\ref{fig:flow_compare}. It can be seen that by using our regularization loss the flow fields are more smooth. Unnecessary jumps are avoided. We use the obtained flow fields to warp the source images directly to show the sampling correctness. It can be seen that without using the regularization loss incorrect sampling regions will be assigned due to the sharp jumps of the flow fields. Relatively good results can be obtained by using the affine assumption prior.





\begin{figure}[b]
\renewcommand\thefigure{\Alph{alphasect}.\arabic{figure}}
\begin{center}
\includegraphics[width=1\linewidth]{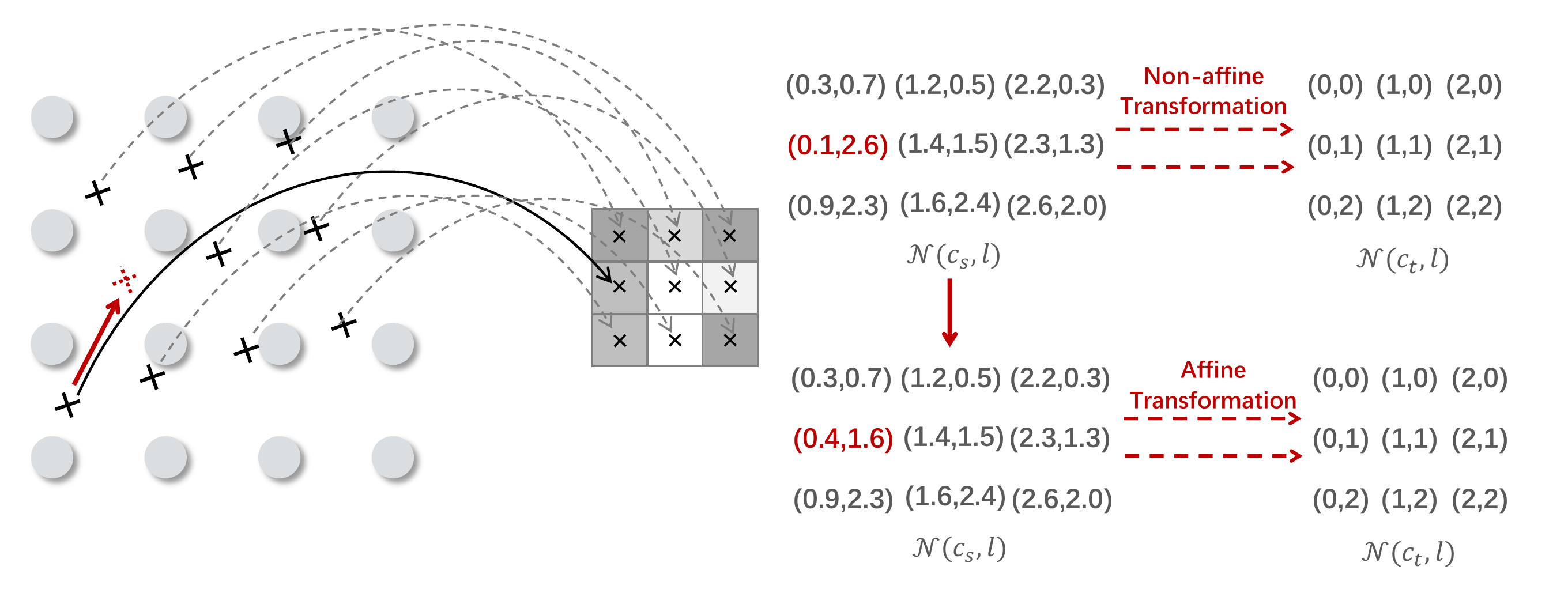}
\end{center}
\caption{Our regularization loss assigns large errors to local regions where the transformation is not an affine transformation thereby forcing the network to change the sampling locations.}
\label{fig:regularization}

\end{figure}

\begin{figure}[b]
\renewcommand\thefigure{\Alph{alphasect}.\arabic{figure}}
\begin{center}
\includegraphics[width=1\linewidth]{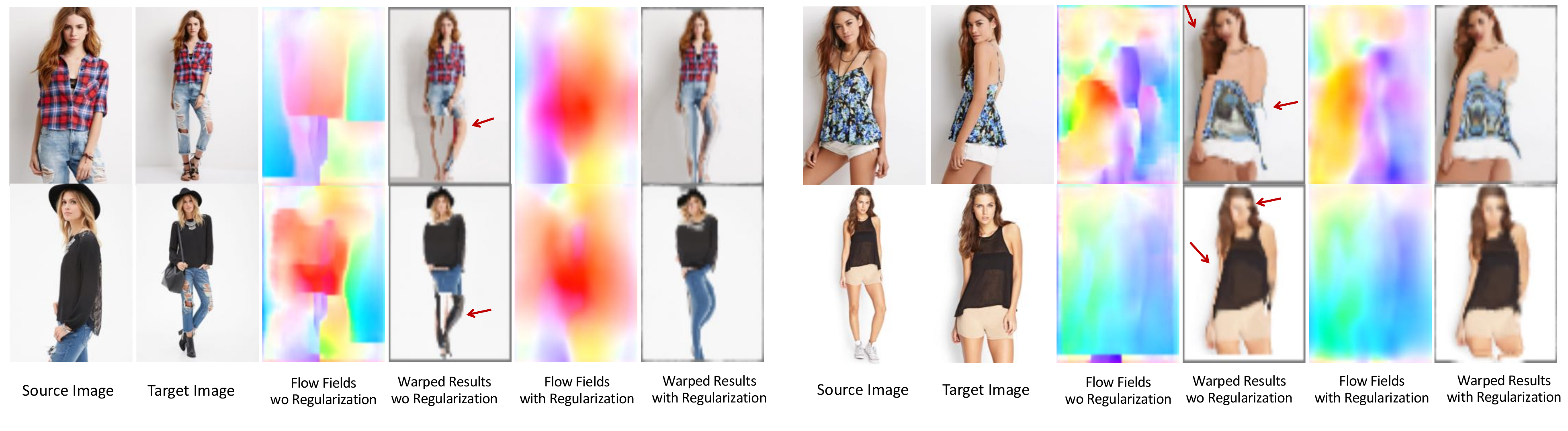}
\end{center}
\caption{The flow fields of models trained with and without the regularization loss. }
\label{fig:flow_compare}
\end{figure}



\section{Additional Results}
\label{sec:additional_result}
\subsection{Additional Comparisons with Existing Works}
We provide additional comparison results in this section. The qualitative results is shown in Figure~\ref{fig:compare}.

\begin{figure}[b]
\renewcommand\thefigure{\Alph{alphasect}.\arabic{figure}}
\begin{center}
\includegraphics[width=0.95\linewidth]{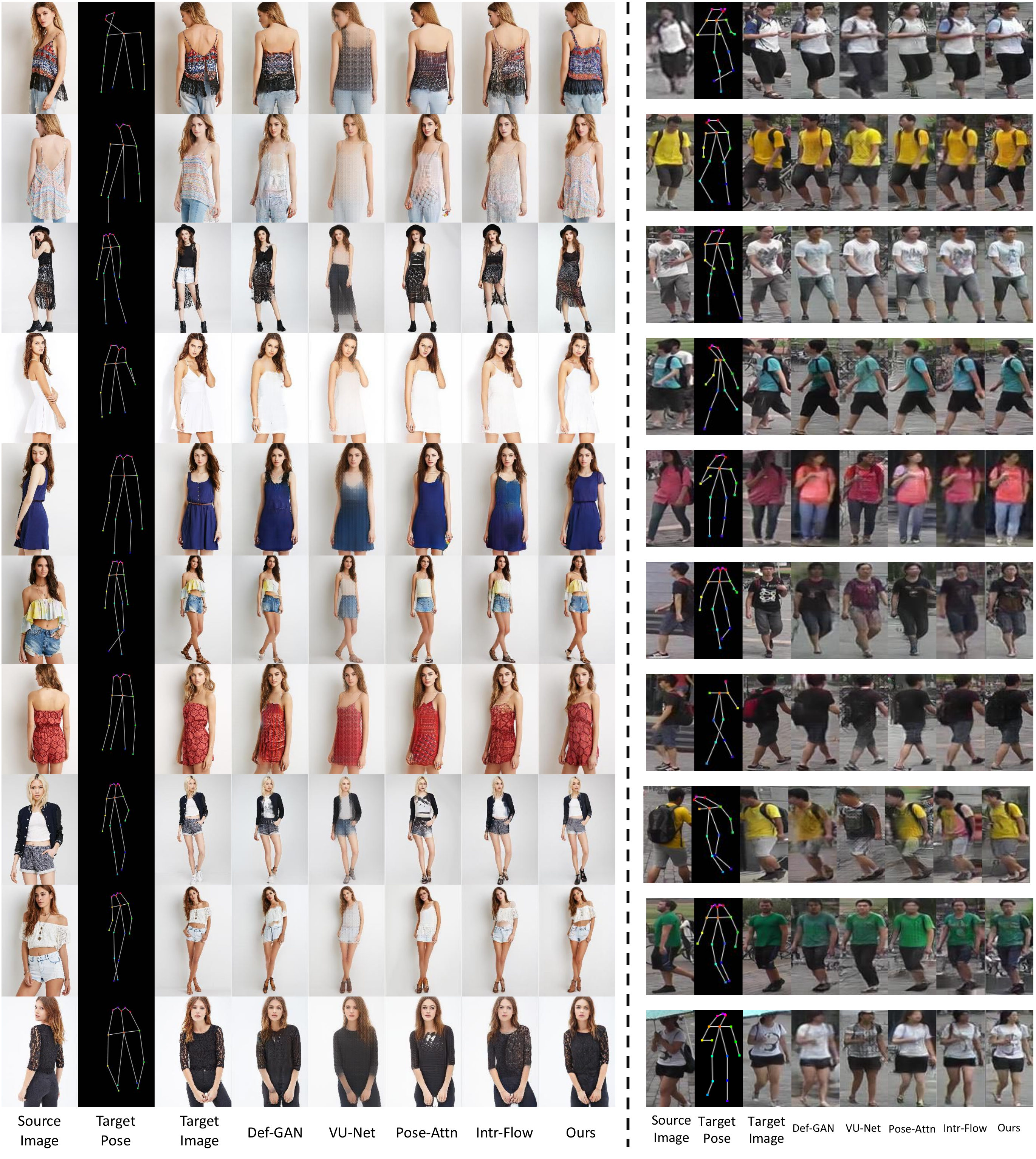}
\end{center}
\caption{The qualitative comparisons with several state-of-the-art models including Def-GAN~\cite{siarohin2018deformable}, VU-Net~\cite{esser2018variational}, Pose-Attn\cite{zhu2019progressive}, and Intr-Flow~\cite{li2019dense} over dataset DeepFashion~\cite{liu2016deepfashion} and Market-1501~\cite{zheng2015scalable}.}
\label{fig:compare}
\end{figure}
\subsection{Additional Results of View Synthesis}
\label{sec:view}
We provide additional results of the view synthesis task in Figure~\ref{fig:car} and Figure~\ref{fig:chair}.
\begin{figure}[b]
    \renewcommand\thefigure{\Alph{alphasect}.\arabic{figure}}
    \offinterlineskip
 
    \centering
    \href{https://user-images.githubusercontent.com/30292465/75650787-f8a81600-5c91-11ea-998e-94685f956302.gif}{
    \includegraphics[width=1\linewidth]{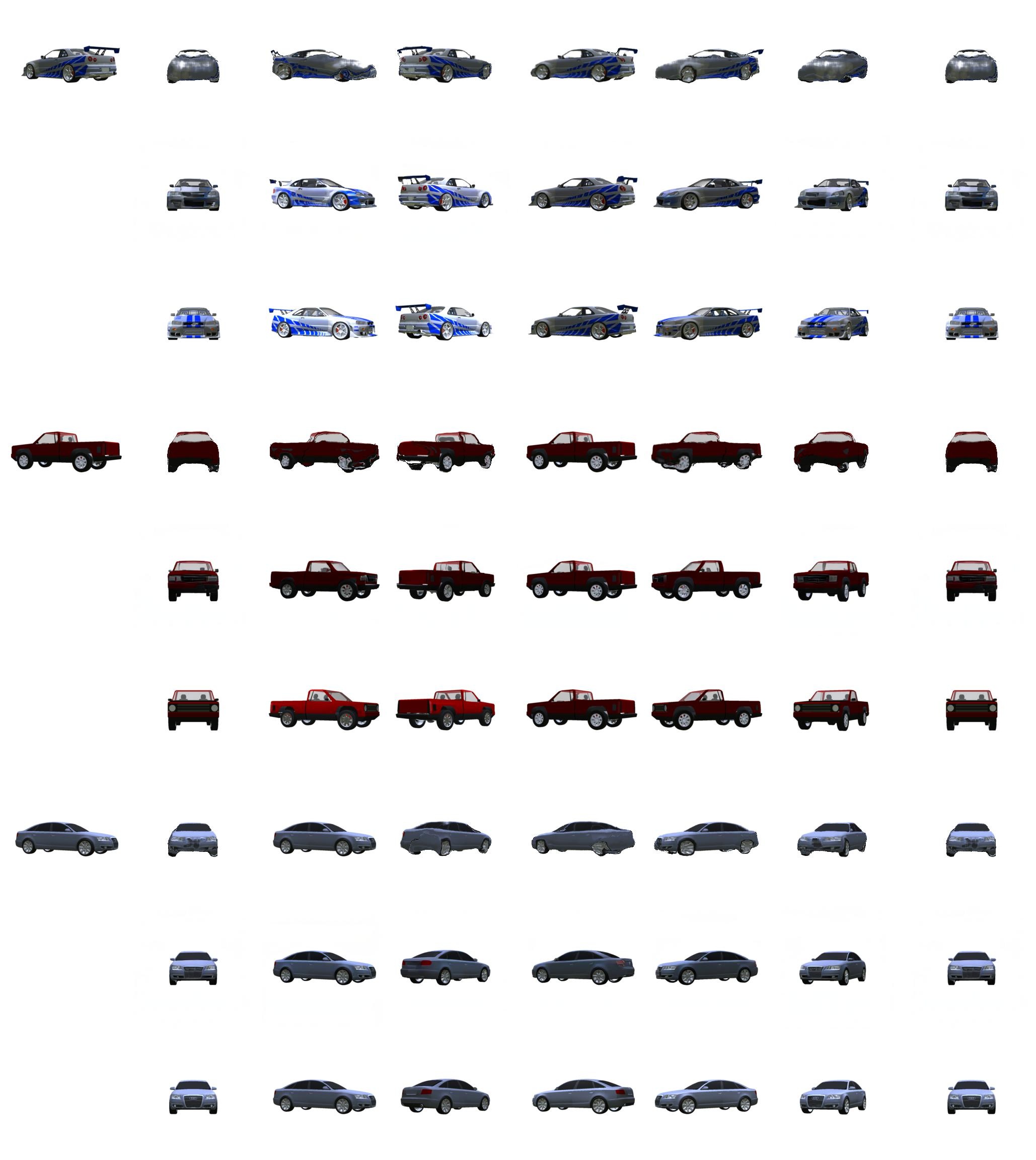}}

    \caption{Qualitative results of the view synthesis task. For each group, we show the results of Appearance Flow~\cite{zhou2016view}, the results of our model, and ground-truth images, respectively. The top left image is the input source image. The other images are the generated results and ground-truth images. \textit{Click on the  image to start the animation in a browser}.}
    \label{fig:car}
\end{figure}

\begin{figure}[b]
    \renewcommand\thefigure{\Alph{alphasect}.\arabic{figure}}

    \offinterlineskip

    \centering
    \href{https://user-images.githubusercontent.com/30292465/75650814-09588c00-5c92-11ea-8c08-52662312c81d.gif}{
    \includegraphics[width=1\linewidth]{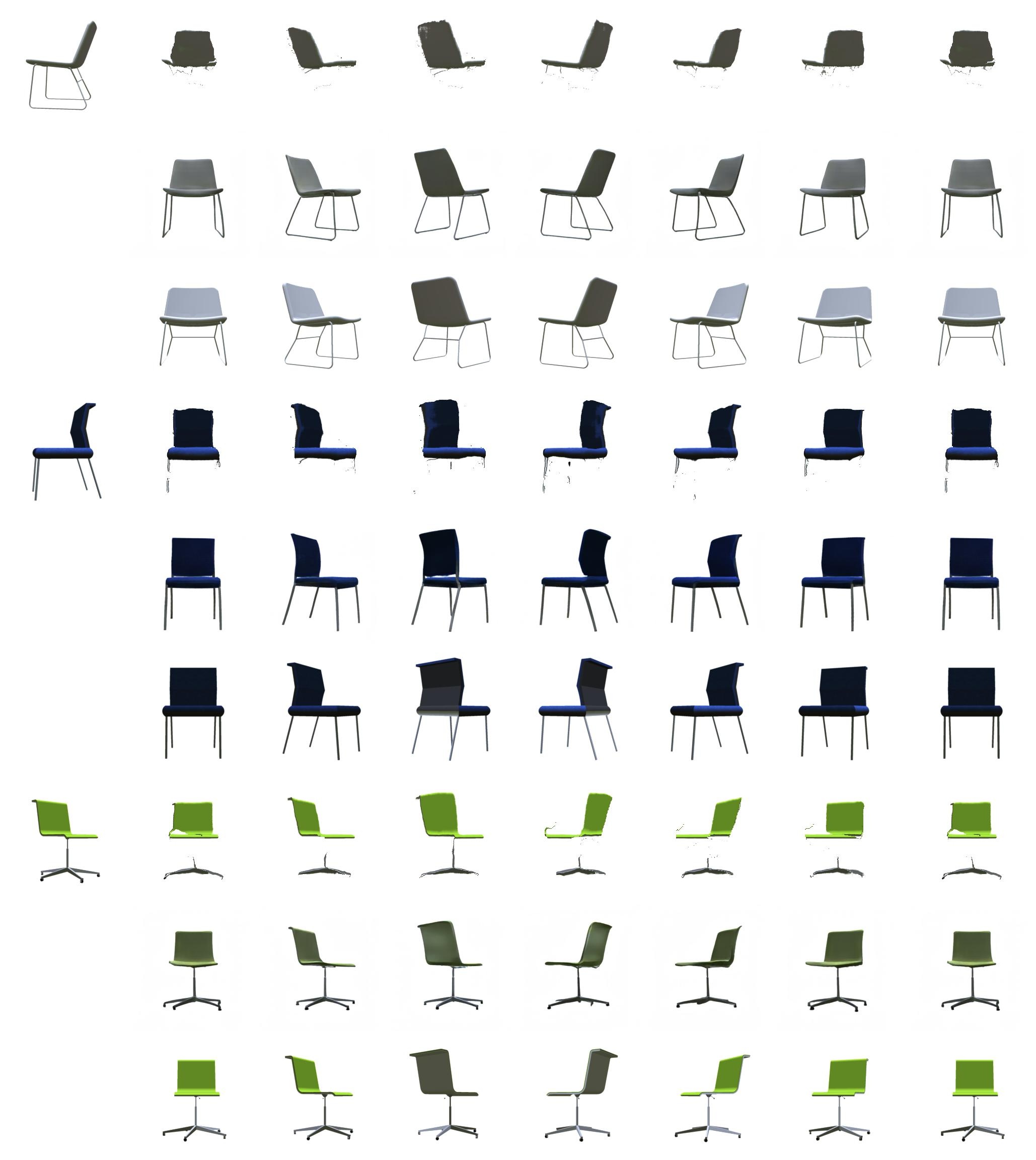}}

    \caption{Qualitative results of the view synthesis task. For each group, we show the results of Appearance Flow~\cite{zhou2016view}, the results of our model, and ground-truth images, respectively. The top left image is the input source image. The other images are the generated results and ground-truth images. \textit{Click on the image to start the animation in a browser}.}
    \label{fig:chair}
\end{figure}

\clearpage

\subsection{Additional Results of Image Animation}
\label{sec:animation}
We provide additional results of the image animation task  in Figure~\ref{fig:face}.
\begin{figure}[ht]
    \renewcommand\thefigure{\Alph{alphasect}.\arabic{figure}}

    \offinterlineskip

    \centering
    \href{https://user-images.githubusercontent.com/30292465/75650849-1d03f280-5c92-11ea-9f3d-ae4c85524787.gif}{
    \includegraphics[width=1\linewidth]{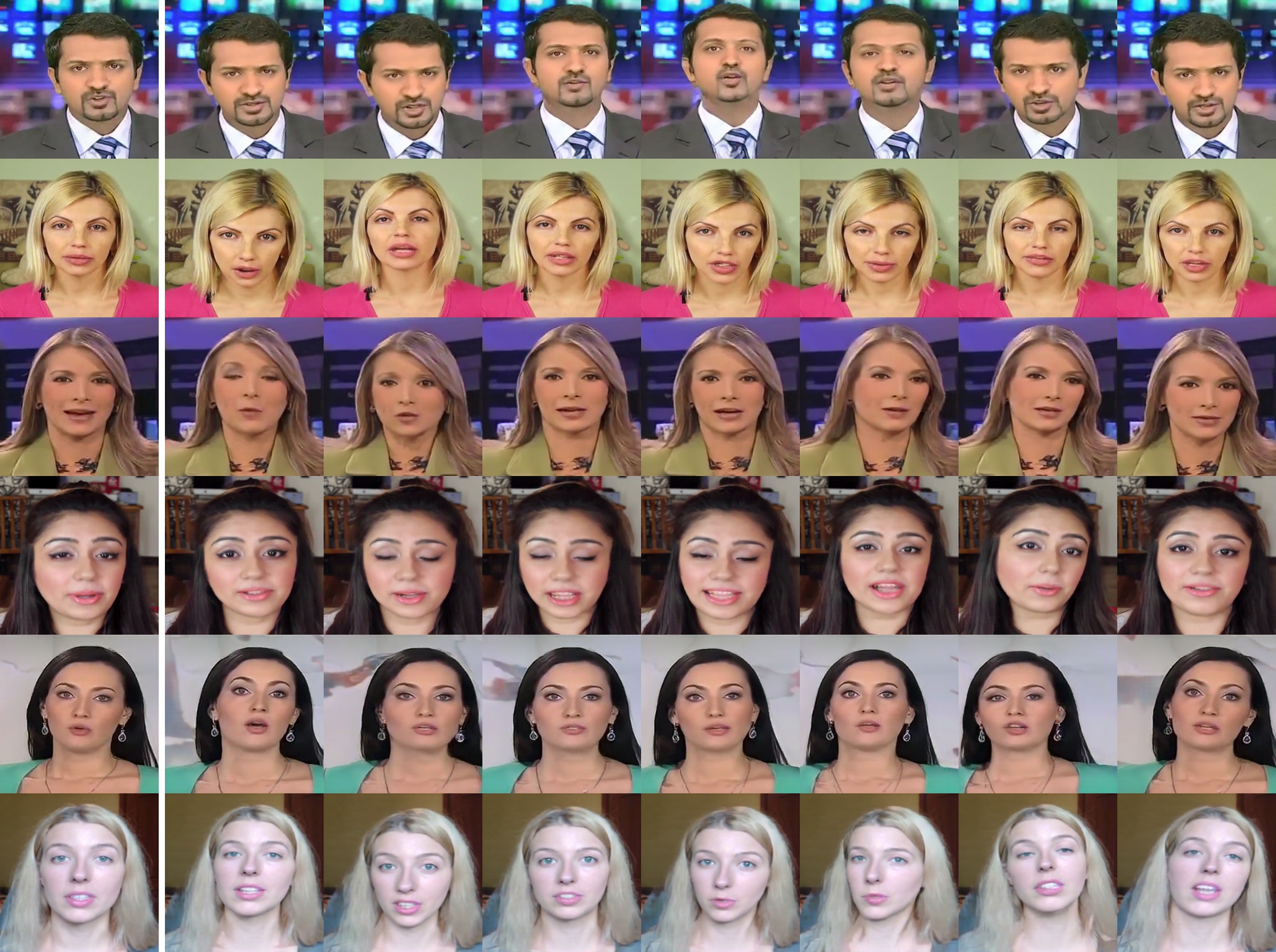}}
 
    \caption{Qualitative results of the image animation task. For each row, the leftmost image is the source image. The others are generated images. \textit{Click on the image to start the animation in a browser}.}
    \label{fig:face}
\end{figure}

\section{Implementation Details}
\label{sec:implementation_detail}
Basically, the auto-encoder structure is employed to design our network. We use the residual blocks as shown in Figure~\ref{fig:network_component} to build our model. Each convolutional layer is followed by instance normalization~\cite{ulyanov2016instance}. We use Leaky-ReLU as the activation function in our model. Spectral normalization~\cite{miyato2018spectral} is employed in the discriminator to solve the notorious problem of instability training of generative adversarial networks. 
The architecture of our model is shown in Figure~\ref{fig:network}. We note that since the images of the Market-1501 dataset are low-resolution images ($128 \times 64$), we only use one local attention block at the feature maps with resolution as $32 \times 16$. 
We design the kernel prediction net $M$ in the local attention block as a fully connected network. The extracted local patch $\mathcal{N}_n(\mathbf{f}_s,l+\mathbf{w}^l)$ and $\mathcal{N}_n(\mathbf{f}_t,l)$ are concatenated as the input. The output of the network is $\mathbf{k}_l$.
Since it needs to predict attention kernels $\mathbf{k}_l$ for all location $l$ in the feature maps, we use a convolutional layer to implement this network, which can take advantage of the parallel computing power of GPUs.  

We train our model in stages. The Flow Field Estimator is first trained to generate flow fields. Then we train the whole model in an end-to-end manner. We adopt the ADAM optimizer. The learning rate of the generator is set to $10^{-4}$. The discriminator is trained with a learning rate of one-tenth of that of the generator. The batch size is set to 8 for all experiments. The loss weights are set to $\lambda_c=5$, $\lambda_r=0.0025$, $\lambda_{\ell_1}=5$, $\lambda_a=2$, $\lambda_p=0.5$, and $\lambda_s=500$. 
\begin{figure}[t]
\renewcommand\thefigure{\Alph{alphasect}.\arabic{figure}}
\begin{center}
\includegraphics[width=0.9\linewidth]{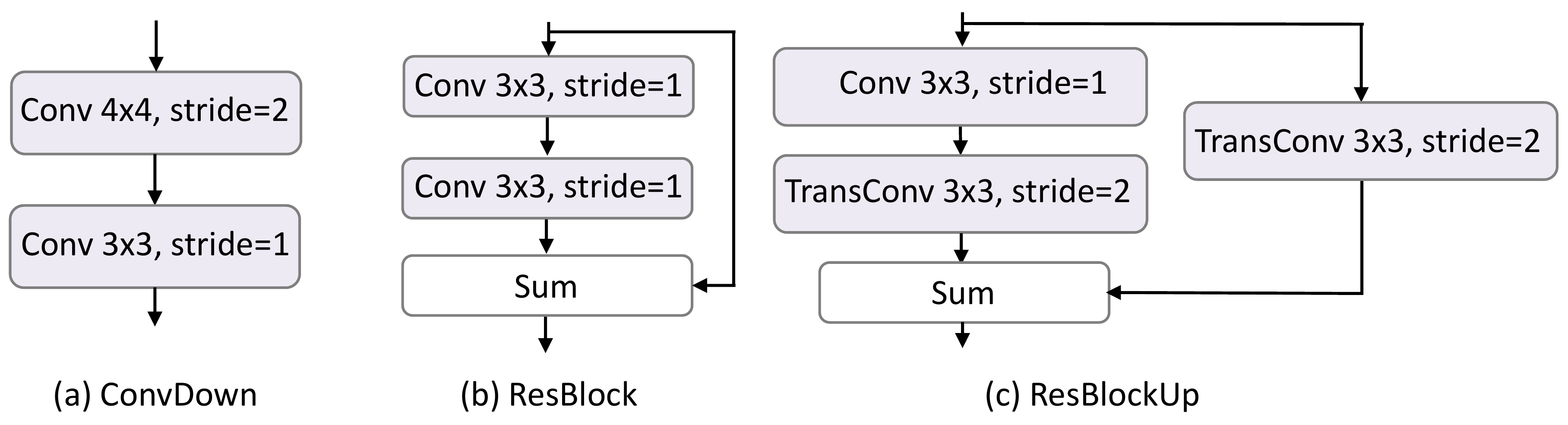}
\end{center}
\caption{The components used in our networks.}
\label{fig:network_component}

\end{figure}

\begin{figure}[b]
\renewcommand\thefigure{\Alph{alphasect}.\arabic{figure}}
\begin{center}
\includegraphics[width=0.9\linewidth]{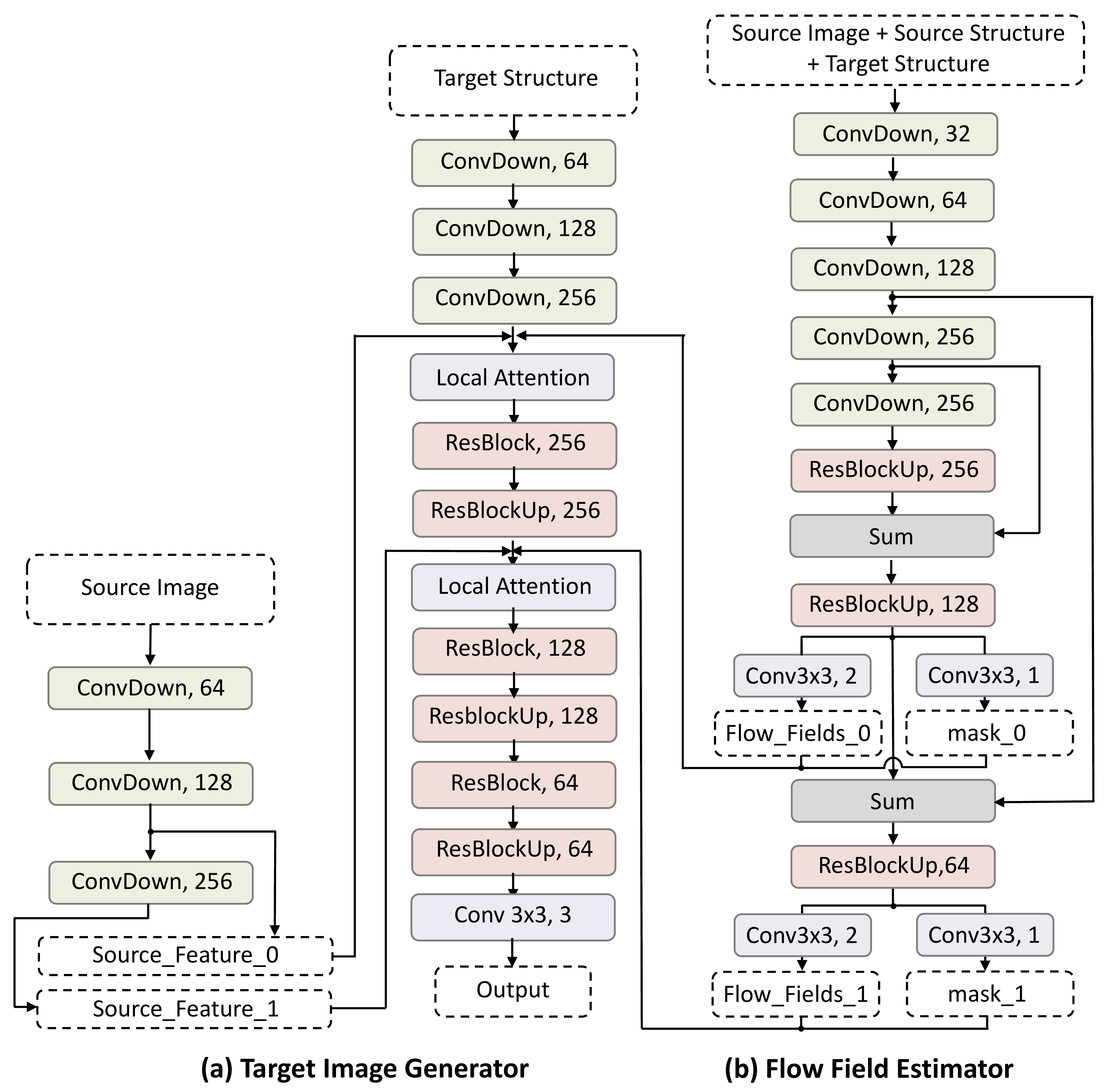}
\end{center}
\caption{The network architecture of our network.}
\label{fig:network}

\end{figure}

\clearpage
\end{alphasection}
\end{document}